\documentclass{article}

\usepackage{microtype}
\usepackage{graphicx}
\usepackage{booktabs} % for professional tables
\usepackage{float}    % For precise placement
\usepackage{subcaption}
\usepackage{natbib}

\usepackage{enumitem}

\usepackage{hyperref}

\usepackage[accepted]{icml2025}
\usepackage{amsmath}
\usepackage{amssymb}
\usepackage{mathtools}
\usepackage{amsthm}

\usepackage[capitalize,noabbrev]{cleveref}

\theoremstyle{plain}

\theoremstyle{definition}

\theoremstyle{remark}

\usepackage[textsize=tiny]{todonotes}

\icmltitlerunning{MatrixKAN: Parallelized Kolmogorov-Arnold Network}

\begin{document}

\twocolumn[
\icmltitle{MatrixKAN: \\
            Parallelized Kolmogorov-Arnold Network}

% It is OKAY to include author information, even for blind
% submissions: the style file will automatically remove it for you
% unless you've provided the [accepted] option to the icml2025
% package.

% List of affiliations: The first argument should be a (short)
% identifier you will use later to specify author affiliations
% Academic affiliations should list Department, University, City, Region, Country
% Industry affiliations should list Company, City, Region, Country

% You can specify symbols, otherwise they are numbered in order.
% Ideally, you should not use this facility. Affiliations will be numbered
% in order of appearance and this is the preferred way.
\icmlsetsymbol{equal}{*}

\begin{icmlauthorlist}
\icmlauthor{Cale Coffman}{yyy}
\icmlauthor{Lizhong Chen}{yyy}
%\icmlauthor{Firstname3 Lastname3}{comp}
%\icmlauthor{Firstname4 Lastname4}{sch}
%\icmlauthor{Firstname5 Lastname5}{yyy}
%\icmlauthor{Firstname6 Lastname6}{sch,yyy,comp}
%\icmlauthor{Firstname7 Lastname7}{comp}
%\icmlauthor{}{sch}
%\icmlauthor{Firstname8 Lastname8}{sch}
%\icmlauthor{Firstname8 Lastname8}{yyy,comp}
%\icmlauthor{}{sch}
%\icmlauthor{}{sch}
\end{icmlauthorlist}

\icmlaffiliation{yyy}{School of Electrical Engineering and Computer Science, Oregon State University, Corvallis, OR, United States.}
%\icmlaffiliation{comp}{Company Name, Location, Country}
%\icmlaffiliation{sch}{School of ZZZ, Institute of WWW, Location, Country}

\icmlcorrespondingauthor{Cale Coffman}{coffmaca@oregonstate.edu}
\icmlcorrespondingauthor{Lizhong Chen}{chenliz@oregonstate.edu}

% You may provide any keywords that you
% find helpful for describing your paper; these are used to populate
% the "keywords" metadata in the PDF but will not be shown in the document
\icmlkeywords{Machine Learning, ICML}

\vskip 0.3in
]

% this must go after the closing bracket ] following \twocolumn[ ...

% This command actually creates the footnote in the first column
% listing the affiliations and the copyright notice.
% The command takes one argument, which is text to display at the start of the footnote.
% The \icmlEqualContribution command is standard text for equal contribution.
% Remove it (just {}) if you do not need this facility.

\printAffiliationsAndNotice{}  % leave blank if no need to mention equal contribution
%\printAffiliationsAndNotice{\icmlEqualContribution} % otherwise use the standard text.

\begin{abstract}

Kolmogorov-Arnold Networks (KAN) are a new class of neural network architecture representing a promising alternative to the Multilayer Perceptron (MLP), demonstrating improved expressiveness and interpretability.  However, KANs suffer from slow training and inference speeds relative to MLPs due in part to the recursive nature of the underlying B-spline calculations.  This issue is particularly apparent with respect to KANs utilizing high-degree B-splines, as the number of required non-parallelizable recursions is proportional to B-spline degree.  We solve this issue by proposing \textit{MatrixKAN}, a novel optimization that parallelizes B-spline calculations with matrix representation and operations, thus significantly improving effective computation time for models utilizing high-degree B-splines.  In this paper, we demonstrate the superior scaling of MatrixKAN's computation time relative to B-spline degree.  Further, our experiments demonstrate speedups of approximately 40x relative to KAN, with significant additional speedup potential for larger datasets or higher spline degrees.
\footnote{
\begin{scriptsize}
\underline{Code}: Our publicly available implementation of MatrixKAN
is provided in the following GitHub repository: \href{https://github.com/OSU-STARLAB/MatrixKAN}{https://github.com/OSU-STARLAB/MatrixKAN}
\end{scriptsize}
}

\end{abstract}

\section{Introduction}
\label{Introduction}

Kolmogorov-Arnold Networks (KAN) are a new class of neural network architecture representing a promising alternative to Multilayer Perceptrons (MLP).  Like MLPs, KANs are comprised of a fully-connected network of nodes and edges.  However, unlike MLPs, which utilize fixed activation functions on network nodes, KANs utilize learnable activation functions on network edges. Along each edge of a KAN, a parameterized B-spline is utilized to model functions of arbitrary degree, making KANs both more expressive and more interpretable \cite{samadi2024}.

Initial empirical testing of KANs has yielded promising results \cite{liu2024, jamali2024, peng2024, hu2024, Koenig2024, decarlo2024, zhou2024, liuchen2025}.  For example, compared to MLPs with comparable parameter counts, KANs trained to model various physics functions have (at worst) achieved comparable loss levels and (at best) achieved loss levels several orders of magnitude better than their MLP counterparts \cite{liu2024}.  While KANs have demonstrated the potential to outperform traditional MLPs, due to their use of parameterized B-splines as learnable activation functions, KANs suffer from significantly slower training and inference speeds.  One major contributing factor to this increase in computation time is the recursive algorithm required for calculating B-spline outputs—the Cox-De Boor recursion algorithm.

As the name suggests, the Cox-De Boor recursion algorithm requires sequential, recursive calls to calculate a single B-spline output, and the number of recursive calls varies linearly with the degree of the B-spline.  Perhaps due to this limitation, KANs typically utilize relatively low-degree B-splines (e.g., degree 3); however, our experiment suggests that certain classes of functions may be more effectively modeled by higher-degree B-splines and that use of higher-degree B-splines may result in enhanced model expressiveness.

To reduce the time complexity of using high-degree B-splines as learnable activation functions, we propose MatrixKAN, a novel optimization that parallelizes B-spline calculations, significantly improving effective computation time for KANs utilizing high-degree B-splines.  This is achieved by adapting the generalized matrix representation of B-splines to KAN and parallelizing the Cox-De Boor recursion algorithm by decomposing its calculations into matrix operations, with all recursive operations represented by a single matrix that is precalculated at model initialization for B-splines of a given degree.  To substantiate this optimization, we perform detailed analyses of the theoretical complexity of each architecture and demonstrate that the optimized MatrixKAN approach results in significantly improved scaling of B-spline calculation time with respect to B-spline degree.

We trained and evaluated our MatrixKAN models on the Feynman physics equations modeled by KAN in \cite{liu2024} and compared both computational efficiency and model performance.  To facilitate comparison with KAN, we evaluated computational efficiency and model performance using the same metrics reported in \cite{liu2024}—(i) seconds per training step and (ii) RMSE loss, respectively.   With respect to computational efficiency, our results demonstrate that KAN's effective computation time scales linearly relative to B-spline degree, while MatrixKAN's effective computation time is independent of B-spline degree, resulting in speedups of up to approximately 40x, with potential for greater speedups using large datasets or higher spline degrees.  With respect to model performance, we demonstrate that MatrixKAN achieves loss levels consistent with those achieved by KAN.  Further, we show that for various modeled functions, KANs of relatively high degree (i.e., 4 or higher) yield improved loss levels, up to approximately 27.0\%, as compared to KANs of relatively low degree (i.e., 3 or less)—cases for which use of the MatrixKAN architecture would result in significantly reduced training and inference time.

The main contributions of this paper are:

\begin{enumerate}[topsep=0pt,itemsep=0ex]
\item Proposed adaptation of the generalized matrix representation of B-splines for B-spline calculations in KAN. 
\item Developed MatrixKAN, an efficient, parallelized implementation of KAN.
\item Demonstrated that the effective computation time of MatrixKAN scales preferably relative to the degree of the B-spline employed within the model, resulting in more efficient calculations for high-degree B-splines.
\item Demonstrated that KANs / MatrixKANs utilizing relatively high-degree B-splines yield improved performance when modeling various functions.
\end{enumerate}

\section{Background and Related Work}
\subsection{Kolmogorov-Arnold Networks}
In addition to the features already noted, KANs differ from MLPs in their theoretical foundation.  Unlike MLPs, whose theoretical basis as a function modeling framework derives from the universal approximation theorem,  KANs rely on the Kolmogorov-Arnold representation theorem, which provides that every multivariate continuous function \(f\) can be represented as a finite composition of continuous, univariate functions and the binary operation of addition:
\begin{equation}
f(x) = f(x_1, ..., x_n) = \sum_{q=1}^{2n+1} \Phi_q (\sum_{p=1}^n \phi_{q,p} (x_p))
\end{equation}
where \(\phi_{q,p} : [0,1] \rightarrow \mathbb{R}\) and \(\Phi_q : \mathbb{R} \rightarrow \mathbb{R}\) \cite{kolmogorov1957, liu2024}. 
This traditional formulation of the Kolmogorov-Arnold representation theorem is analogous to a small neural network with (i) \(n\) input neurons, (ii) a single hidden layer with \(2n+1\) neurons, and (iii) a single output neuron.  \(\phi_{q,p}\) represent the edges between input neurons and neurons in the hidden layer, and \(\Phi_q\) represent edges between neurons in the hidden layer and the output neuron.

To generalize this approach for modeling more complex systems and to facilitate training of this using traditional neural network regression strategies, KANs expand this two-layer network to networks of arbitrary width and depth and parameterize all edge functions using B-splines.  The shape of a KAN is represented as \([n_0, n_1, ..., n_L]\), where \(n_i\) is the number of nodes in the \(i^{th}\) layer.  The activation value for a given node is represented as \(x_{l,i}\) and the B-spline activation function between two nodes in adjacent layers is represented as \(\Phi_{l, j, i}\), where \(l\) denotes the layer, \(i\) denotes the neuron within the given layer and \(j\) denotes the connected neuron in the subsequent layer \cite{liu2024}:
\begin{equation}
\phi_{l,j,i}, l=0, ..., L-1, i=1, ..., n_l, j=1, ..., n_{l+1}
\end{equation}

The output of a given B-spline activation function is represented as \(\tilde{x} \equiv \Phi_{l,j,i} (x_{l,i})\).   For any given node, the activation value is equal to the sum of the outputs of the B-spline activation functions on each incoming network edge \cite{liu2024}:
\begin{equation}
x_{l+1, j} = \sum_{i=1}^{n_l} \tilde{x}_{l,j,i} = \sum_{i=1}^{n_l} \phi_{l,j,i} (x_{l,i})
\end{equation}
where \(j = 1, ..., n_{l+1}\).

\subsection{B-Splines}
\begin{figure} [ht]
\begin{center}
\centerline{\includegraphics[width=0.75\columnwidth]{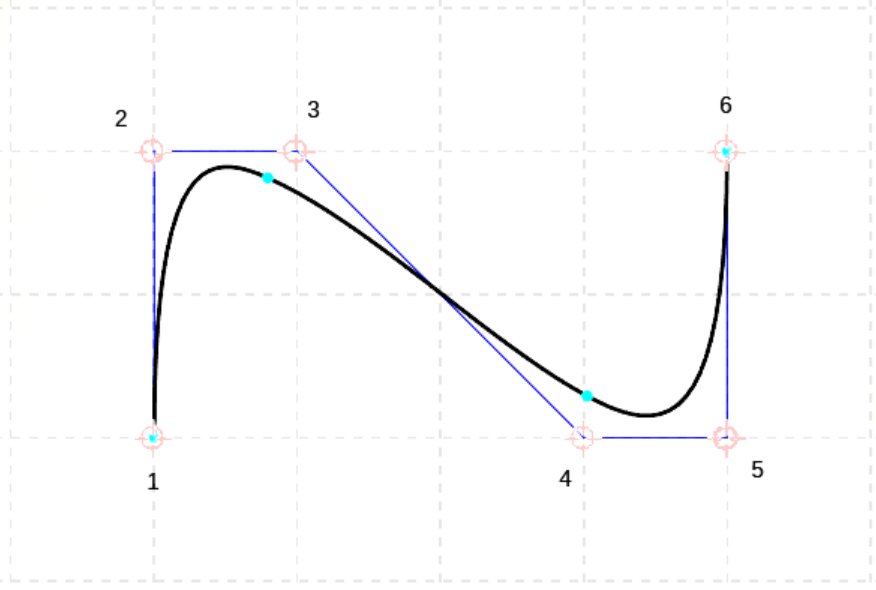}}
\caption{Diagram of a cubic B-spline curve, including applicable knots (turquoise) and control points (red).}
\label{fig:b-spline curve}
\end{center}
\end{figure}
To facilitate diversity in the learned activation function on each edge of a KAN, KANs implement activation functions using B-splines.  B-splines are piecewise polynomial curves whose name derives from the elastic beams (i.e., splines) used by draftsmen to construct sweeping curves in ship design \cite{prautzsch2002}.  They are defined as affine combinations of control points \(c_i\) and associated basis functions \(B_i (x)\) \cite{liu2024}:
\begin{equation} \label{eq:KANspline-1}
spline(x) = \sum_i c_i B_i(x)
\end{equation}

\begin{figure} [ht]
\begin{center}
\centerline{\includegraphics[width=0.75\columnwidth]{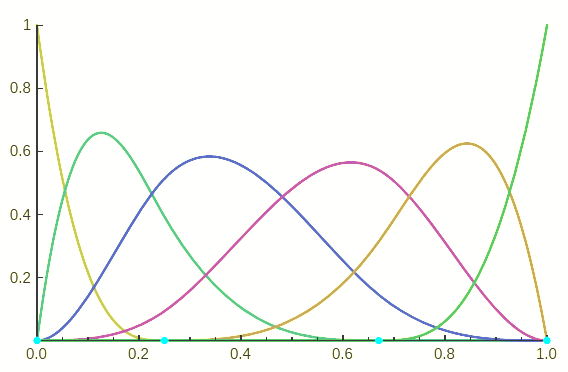}}
\caption{Diagram of cubic B-spline basis function curves (various colors) and knots (turquoise).}
\label{fig:b-spline basis functions}
\end{center}
\end{figure}

B-splines can be used to model a wide variety of shapes due to their ability to represent functions of arbitrary degree.  A curve \(spline(x)\) is called a B-spline of degree \(k\) with knots \(t_o, ..., t_m\), where \(t_i \leq t_{i+1}\) and \(t_i < t_{i+k+1}\) for all \(i\), which delineate the intervals over which the B-spline is defined (see Figures \ref{fig:b-spline curve} and \ref{fig:b-spline basis functions}), and control points \(c_o, ..., c_i\), which control the local shape of the B-spline. Due to their control over the shape of the B-spline, KANs parameterize B-splines via their control points. Expressions for the basis functions \(B_{i,k} (x)\) of a given B-spline are derived via the Cox-De Boor recursion algorithm, examples of which are shown in Figure \ref{fig:b-spline basis functions}:

\begin{small}
\begin{equation} \label{eq:Cox-De Boor}
\begin{split}
B_{i,0} (x) & =
\begin{cases}
1 & \text{if } t_i \leq x < t_{i+1},\\
0 & \text{otherwise.}
\end{cases} \\
B_{i,k} (x) & = \frac{x - t_i}{t_{i+k} - t_i} B_{i, k-1} (x) + \frac{t_{i+k+1} - x}{t_{i+k+1} - t_{i+1}} B_{i+1, k-1} (x)
\end{split}
\end{equation}
\end{small}

As Equations \ref{eq:KANspline-1} and \ref{eq:Cox-De Boor} demonstrate, the value of a basis function \(B_{i,k} (x)\) of degree \(k\) is calculated recursively by reference to the position of the input value \(x\) within the knot vector and the output of certain basis functions of degree \(k-1\), and the final output of the B-spline is determined by multiplying all applicable basis functions by their associated control points \(c_i\).

Given that basis functions of degree \(k\) are defined by reference to basis functions of degree \(k - 1\), it follows that calculation of a degree \(k\) basis function requires \(k\) levels of sequential, recursive calls. This sequential dependency in the Cox-De Boor recursion algorithm prevents parallelization, resulting in a proportional increase in computation time as B-spline degree increases, even with memorization or dynamic programming techniques. Our testing shows that B-spline calculations constitute a significant portion of total execution time for KANs, ranging from 18\% for degree-2 B-splines to over 50\% for degree-30 B-splines, making B-spline computational efficiency a fundamental concern in improving the computational efficiency of KANs as a whole. While low-degree B-splines enable KANs to outperform MLPs \cite{liu2024, peng2024, hu2024}, our tests have demonstrated that high-degree B-splines are necessary for optimal performance in modeling certain functions, with optimal convergence occurring at degrees of 30 or higher. 
To bridge the gap between the need for high-degree B-splines and their high computational cost, we need a way to parallelize the recursive B-spline calculations for efficient execution on a parallel processor.

\subsection{Related Work}

To address the computational cost of KANs, a number of approaches have been taken.  For certain applications, despite the general additional computational overhead of KANs as compared to MLPs with similar parameter counts, KANs have been shown to produce superior performance with lower parameter counts, resulting in overall computational efficiency, as is the case with the convolutional KAN and U-KAN \cite{bodner2024, li2024}.  However, these results involve hybrid KAN architectures and are domain-specific, suggesting the need for more general optimization techniques.  To achieve more general computational efficiency, many modified KAN architectures have been proposed that replace underlying B-splines with different basis functions, such as radial basis functions (FastKAN / FasterKAN) \cite{li2024radial}, trigonometric functions (e.g., Larctan-SKAN) and left-shift softplus functions (LSS-SKAN) \cite{chen2024lss, chen2024larctan}, and rational functions (rKAN) \cite{aghaei2024}.  While each of these approaches has shown speedups over the B-spline-based KAN architecture, each basis function type has its own strengths and weaknesses in function approximation, and none identified here exhibit the local support feature of B-splines that allows for its granular function modeling capabilities. In certain research, adding wavelet functions to B-spline-based KANs (Wav-KAN) \cite{bozorgasl2024} or implementing neuron grouping / weight sharing to minimize the overall parameter count of B-spline-based KANs (Free Knots KAN) \cite{zheng2025} has yielded efficiency gains, but in all cases these speedups are only by a constant factor and none resolves the underlying computational complexity of the recursive B-spline calculations that is addressed by our proposed approach.  That said, each is complementary to our approach and could be used in combination to achieve further improved computational efficiency.

\section{MatrixKAN}
In this work, we propose \textit{MatrixKAN}, a novel optimization that parallelizes B-spline calculations, allowing for significant reduction in the effective computation time of KAN calculations.  In what follows, we first demonstrate that the generalized matrix representation of B-splines can be theoretically applied to B-spline-based KANs.  Next, we demonstrate the procedure for parallelizing B-spline outputs in MatrixKAN.  Finally, we discuss the computational efficiency of KAN and MatrixKAN.

\subsection{Applying Generalized Matrix Representation of B-Splines to KAN}

\subsubsection{Decomposing the Cox-De Boor Algorithm into Matrix Operations}

We first show that the summation of the products of (i) control points and (ii) basis functions representing the output of a B-spline in KAN (as presented in Equation \ref{eq:KANspline-1}) can be decomposed into a summation of only matrix multiplications--the general matrix representation of B-splines.

For a B-spline of order \(k\)\footnote{Note that a B-spline of order \(k\) is equivalent to a B-spline of degree \(k-1\)},  whose shape is determined by \(k\) control points \(c_j (j=i, ..., i+k-1)\), the output of the \(i^{th}\) B-spline curve segment \(spline_i (u)\) can be represented as follows:
\begin{equation} \label{eq:KANspline-2}
spline_i (u) = \sum_{j=i}^{i+k-1} c_j B_{j,k} (u)
\end{equation}
where \(u = \frac{t - t_{i+k-1}}{t_{i+k} - t_{i+k-1}}, t_{i+k-1} \leq t < t_{i+k}\).

By expanding the summation, we obtain the following:
\begin{small}
\begin{equation}
spline_i (u) = c_i B_{i,k} (u) + c_{i+1} B_{i+1,k} (u) + ... + c_{i+k-1} B_{i+k-1,k} (u)
\end{equation}
\end{small}

This expanded summation can be represented in matrix form as the product of (i) all applicable basis functions and (ii) the control points, as follows:

\begin{tiny}
\begin{equation} \label{eq:gen-bspline-2}
spline_i (u) =
\begin{bmatrix} B_{i,k} (u) & B_{i+1,k} (u) & ... & B_{i+k-1,k} (u) \end{bmatrix}
\begin{bmatrix} c_i \\ c_{i+1} \\ ... \\ c_{i+k-1} \end{bmatrix}
\end{equation}
\end{tiny}

As demonstrated by \cite{cohen1982, qin1998}, because all basis functions \(B_{i,k} (u)\) will be, by definition, degree  \(k\), we can leverage Toeplitz matrices\footnote{For a detailed discussion of the use of Toeplitz matrices to represent the product of polynomials, see Appendix \ref{Toeplitz}.} to further decompose the basis function matrix from Equation \ref{eq:gen-bspline-2} \(\begin{bmatrix} B_{i,k} (u) & B_{i+1,k} (u) & ... & B_{i+k-1,k} (u) \end{bmatrix}\) into two matrices—(i) the power bases \(\begin{bmatrix} 1 & u & ... & u^{k-1} \end{bmatrix}\), which represents the \(k\) degrees of the input value \(u,\) and (ii) the basis matrix \(\Psi^k (i)\):
\begin{equation} \label{eq:matrix spline}
spline_i (u) =
\begin{bmatrix} 1 & u & ... & u^{k-1} \end{bmatrix}
\Psi^k (i)
\begin{bmatrix} c_i \\ c_{i+1} \\ ... \\ c_{i+k-1} \end{bmatrix}
\end{equation}
where \(u = \frac{t - t_{i+k-1}}{t_{i+k} - t_{i+k-1}}, t_{i+k-1} \leq t < t_{i+k}\).

\subsubsection{Precalculating the Basis Matrix}

We now demonstrate that the basis matrix \(\Psi^k\) is identical for all B-splines of a given degree and, as such, can be precalculated such that all subsequent B-spline calculations utilizing the basis matrix are simply matrix multiplications, which can be parallelized.

As demonstrated in \cite{qin1998}, \(\Psi^k (i)\) is calculated via the following recursive formula for all uniform B-splines:

\begin{small}
\begin{equation} \label{eq:basis matrix}
\begin{cases}
\Psi^1 = [1] \\
\Psi^k = \frac{1}{k-1}
    \begin{bmatrix}
        \Psi^{k-1} \\ 0
    \end{bmatrix}
    \begin{bmatrix}
        1 & k-2 &  &  & 0 \\
         & 2 & k-3 &  & \\
         &  & \ddots & \ddots & \\
         0 &  &  &  -1 & 1
    \end{bmatrix}
    + \\
    \;\;\;\;\;\;\;\;\;\;\;\;\;\;\;\;\;\;\;\;\;
    \begin{bmatrix}
        0 \\ \Psi^{k-1}
    \end{bmatrix}
    \begin{bmatrix}
        -1 & 1 &  &  & 0 \\
         & -1 & 1 &  & \\
         &  & \ddots & \ddots & \\
         0 &  &  &  -1 & 1
    \end{bmatrix}
\end{cases}
\end{equation}
\end{small}

Note that because the only input into Equation \ref{eq:basis matrix} is the applicable order \(k\) of the B-spline, the output of this algorithm is always the same \((k-1)\)  x \((k-1)\) matrix for a B-spline of order \(k\), meaning \(\Psi^k (i)\) can be calculated only once for all B-splines of order \(k\). The output of all B-splines of order \(k\) can then be calculated via the matrix multiplication operations set forth in Equation \ref{eq:matrix spline}, without the need to execute any further recursive calculations in accordance with Equations \ref{eq:KANspline-1}, \ref{eq:Cox-De Boor}, and \ref{eq:KANspline-2}.

\subsection{Procedure for Parallelizing B-Spline Outputs in MatrixKAN}

Given the above theoretical analysis, we now present the procedure for precalculating the basis matrix \(\Psi^{k+1}(i)\) of order \(k+1\) (i.e., of degree \(k\)) and executing all B-spline calculations in a MatrixKAN network in a fully parallelized manner:

\textbf{Step 1: Calculate the basis matrix \(\Psi^k\).} At model initialization, calculate the basis matrix applicable to B-splines of the degree, \(k\), specified for the model.  This results in a \(k\) x \(k\) tensor.

\textbf{Step 2: Calculate applicable knot intervals and power bases \(\begin{bmatrix} 1 & u & ... & u^{k} \end{bmatrix}\)}, where \(u = \frac{t - t_{i+k}}{t_{i+k+1} - t_{i+k}}, t_{i+k} \leq t < t_{i+k+1}\) and \(t_o, ..., t_m\) represents the knot vector\footnote{Equation \ref{eq:matrix spline} is described in terms of B-spline order, while the formula here is described in terms of B-spline degree.}.  Thus, for each B-spline, we must first calculate \(u\), which requires determining the applicable knot interval for a given input value.  This is achieved by comparing the input values to the knot values in each B-spline knot vector, returning two tensors—one representing the lower knot in the applicable grid interval for each value \(t_{i+k}\) and one representing the upper knot in the applicable grid interval for each value \(t_{i+k+1}\).  These tensors are then used, along with the tensor containing the input values \(t\), to calculate \(u\), per the formula above.  Once \(u\) is calculated, the power bases tensors is then populated with the appropriate powers of \(u\).

\textbf{Step 3: Calculate applicable basis function outputs.}  Next, we calculate the basis function outputs for each B-spline based on Equation \ref{eq:matrix spline} by multiplying the basis matrix tensor calculated in Step 1 with the power bases tensor calculated in Step 2. 
To ensure that only applicable basis function outputs are calculated, we utilize the applicable knot intervals calculated in Step 2 to zero portions of the basis matrix tensor representing inapplicable basis functions.  We then execute the final matrix multiplication, which results in a tensor containing the basis function outputs for all B-splines across all input values\footnote{The basis matrix tensor may be expanded to match the dimension of the power bases tensor, which has additional dimensions corresponding to the number of input values and number of dimensions of each input value.}.

\textbf{Step 4: Calculate B-spline outputs.}  At this point, the output of Step 3 has completed the parallelized calculation of all equivalent recursive calculations required by the Cox-De Boor recursion algorithm and is equivalent to the basis function output tensor produced by KAN, so we can employ the original KAN algorithm from here forward to complete the B-spline output calculations.

All of the above operations are tensor operations implemented in PyTorch, and unlike the original KAN implementation, the recursive operations required to calculate basis function outputs have been replaced by matrix multiplication operations, which can be effectively parallelized.

\subsection{Computational Efficiency Analysis} \label{sect:comp_eff}

In this section, we analyze and compare the computational efficiency of the algorithms implemented in KAN and MatrixKAN.  For purposes of this discussion, we assume a network that mirrors what is presented in \cite{liu2024}, where all $L$ layers have equal width \(n_0 = n_1 = ... = n_L = N\) with each B-spline of degree \(k\) on \(G\) intervals.

\subsubsection{KAN}

We first demonstrate that (i) the number of computations required by KAN in a forward pass is \(O(N^2 L(k^2+kG)\) and (ii) the effective computation time of KAN when executed on a parallel processor is \(O(Lk)\).

To determine the number of calculations required for a forward pass of KAN with the parameters described above, note there are \(L\) layers and \(N^2\) edges between layers (i.e., \(N^2\) B-splines between layers), resulting in \(O(N^2 L)\) total B-splines. For each B-spline, the recursion algorithm in Equation \ref{eq:Cox-De Boor} can be efficiently implemented with memorization or dynamic programming that iterates over $i$ and $k$ dimensions in a bottom-up fashion, i.e., computing $B_{i,0}$ for all $i$ first, then $B_{i,1}$ for all $i$, and so on. This requires $O(Dim_i \times Dim_k)$ = $O((k+G)k)$ = $O(k^2 + kG)$ calculations.  Thus, each forward pass of KAN entails \(O(N^2 L(k^2+kG))\) calculations.

To determine the effective computation time of a forward pass of KAN, we must consider the distinction between the serial and parallel portions of the algorithm.  As discussed in \cite{blelloch1996}, 
when analyzing the effective computation time of parallel algorithms, because all calculations that have no sequential dependencies can be processed in parallel (assuming a sufficient number of parallel processing cores to spread the work), it is only the length of the longest sequential dependency (i.e., the number of calculations that must be performed sequentially) that governs the effective computation time.

Applying this framework to KAN, first note that the \(O(N^2)\) computations referenced above (i.e., the number of B-spline calculations within a given layer) can be fully parallelized, as the calculations at any layer are only sequentially dependent upon the outputs of the prior layer.  Thus, with respect to the B-spline calculations across the network, the longest chain of sequential dependencies is equal to the number of layers, or \(O(L)\).  Further, as discussed above, although each B-spline has \(O(k^2 + kG)\) complexity, the computations at each of the $k$ levels of recursion (e.g., computing $B_{i,0}$ for all $i$) are independent of each other and only sequentially dependent upon the prior level of recursion. Therefore,the longest chain of sequential dependencies for B-spline calculations in a given layer is equal to the total levels of recursion, or \(O(k)\).  When multiplied with the number of layers, \(O(L)\), this results in \textbf{the total effective computation time for a KAN forward pass of $(O(Lk)$}.

\subsubsection{MatrixKAN}

In this section, we demonstrate that (i) the number of computations required by MatrixKAN in a forward pass is \(O(N^2L(k^2 + G)\) and (ii) the effective computation time of MatrixKAN when executed on a parallel processor is superior to that of KAN--\(O(L)\).

First, we address the number of computations required by MatrixKAN in precalculating the basis matrix \(\Psi^{k+1} (i)\).  As set forth in Equation \ref{eq:basis matrix}, calculation of \(\Psi^{k+1} (i)\) requires \(k\) levels of recursion, each of which entails two sequentially independent matrix multiplications of square matrices and one addition of the resultant matrices, starting with matrices of dimension \(k\) x \(k\) and decrementing by \(1\) with each recursive call down to \(1\) x \(1\).  Thus, each step requires \(O(2d^3+d^2)\), or simply \(O(d^3)\), calculations, where \(d\) represents the matrix dimension at each level of recursion.  This can be represented by the following summation, requiring \(O(k^4)\) computations:
\begin{equation}
\sum_{d=1}^k d^3 = \frac{k^2 (k+1)^2}{4} = O(k^4)
\end{equation}

Here, to determine the effective computation time of calculating the basis matrix \(\Psi^{k+1} (i)\), note that because all \(O(d^3)\) calculations required at each level of recursion are matrix operations, all are sequentially independent and can be parallelized.  Thus, the longest chain of sequential dependencies is the total levels of recursion, making the effective computation time of the basis matrix \(\Psi^{k+1} (i)\) equal to \(O(k)\).  Note further that because this calculation need only be executed once for a given B-spline degree (i.e., potentially only once \textit{ever}, as the calculated basis matrix can be reused between models), the amortized computation time of the algorithm is \(O(k^4/n)\), where \(n\) represents the total number of forward passes executed.  Given that model training can easily entail thousands (if not orders of magnitude more) forward passes, \(n\) will eventually grow very large, and the amortized computation time of calculating the basis matrix \(\Psi^k (i)\) will eventually approach \(O(1)\).

Next, we address the number of computations required by MatrixKAN in a forward pass.  Here, as with KAN, there are  \(L\) layers and \(N^2\) B-splines between layers, resulting in \(O(N^2 L)\) total B-splines.  
Each of the B-splines is calculated based on Equation \ref{eq:matrix spline}, which requires calculation of the power bases matrix $u$, which entails $O(G + k)$ complexity due to the dimension of the knot vectors, and two matrix multiplications.  The first matrix multiplication is between the \(1\) x \(k\) power bases matrix  and the \(k\) x \(k\) basis matrix, requiring \(O(k^2)\) computations, and the second is between the \(1\) x \(k\) matrix resulting from the first calculation and the \(k\) x \(1\) control points matrix, requiring  \(O(k)\) computations.   Consequently, each B-spline calculation in a given layer is \(O(G + k + k^2 + k)\), or simply \(O(k^2 + G)\). This, when multiplied with \(O(N^2L)\) total B-splines, results in a total number of \(O(N^2L(k^2 + G))\) calculations required in a forward pass of MatrixKAN.

We now demonstrate how the enhanced parallelizability of MatrixKAN B-spline calculations results in significantly reduced effective computation time.  To determine the effective computation time of a forward pass of MatrixKAN, first note that, as with KAN, all \(O(N^2)\) B-spline computations in a given layer can be parallelized.  Also like KAN, calculations at any given layer are sequentially dependent upon calculations at the prior layer, resulting in a chain of \(O(L)\) sequential dependencies across layers of the network.  However, unlike KAN, all B-spline calculations within a given layer \textit{can} be fully parallelized, i.e., the aforementioned $O(k^2 + G)$ matrix-based calculations in Equation \ref{eq:matrix spline} can be processed in parallel in $O(1)$ time.
Thus, the only chain of sequentially dependent calculations for a given MatrixKAN forward pass are the calculations performed across layers, \textbf{resulting in an effective computation time for MatrixKAN of \(O(L)\)}. 

To summarize, in terms of computational complexity, MatrixKAN requires $O(N^2L(k^2+G))$ computations, which is less than the $O(N^2L(k^2+kG))$ computations required by KAN. More importantly, in terms of effective computation time, MatrixKAN's \(O(L)\) complexity is $k$ times faster than KAN's $O(Lk)$ complexity, which represents a speedup proportional to the degree of the underlying B-spline.  As demonstrated in Section \ref{sect:results}, for networks utilizing relatively high-degree B-splines, this can result in calculations many multiples, if not orders of magnitude, faster than KAN.

\section{Experimental Setup}
\subsection{Data Sets}

Following the evaluation methodology presented in \cite{liu2024}, we conducted experiments on (i) the dataset generated by the hellokan.ipynp script (for the efficiency tests) and (ii) a random subset of the Feynman equations generated by the feynman.py script (for the performance tests), each of which is provided by \cite{pykan}.   For each test, we generated a training set containing 1000 training samples and a test set containing 1000 validation samples. After each training epoch, each model was validated against the test set.

\subsection{Computational Efficiency}

For the efficiency tests, our aim was to compare the computation time of KAN\footnote{For all testing, we utilized the latest MultKAN implementation of KAN provided by \cite{pykan}.} and MatrixKAN relative to grid size, B-spline degree, and dataset size.  We trained models using one NVIDIA 1080ti-8MB GPU for 20 steps, and to mirror the metrics in \cite{liu2024}, we measured computational efficiency in seconds per step, averaged across all steps.  With this general setup, we ran three tests comparing--(1) training time vs. grid size, (2) training time vs. spline degree, and (3) training time vs. dataset size.

\subsection{Performance / Optimal Spline Degree}

For performance tests, our aim was to demonstrate (i) the validity of MatrixKAN, by demonstrating model performance consistent with that of KAN and (ii) the superior performance of models using high-degree B-splines (i.e., of degree 4 or higher) relative to models using low-degree B-splines (i.e. of degree 3 or less) for certain modeling tasks.  We trained KAN and MatrixKAN models in tandem using the shapes prescribed in \cite{liu2024} Table 2 Feynman dataset.
To ensure comparability between KAN and MatrixKAN models, models of each architecture were trained using identical training and test sample sets and identical random seed values.   Training was performed using the Adam optimizer with a learning rate of 0.001.  We trained models of each architecture for 1000 steps using the following values for the underlying B-spline degree: 2, 4, 6, 8, 10, 20, 30.

\section{Results} \label{sect:results}
\subsection{Computational Efficiency}

To compare the computational efficiency of MatrixKAN to KAN, we trained models of shape [2,2,1] and [2,5,1] using the LBFGS optimizer, each model initialized with a grid size of 3, a grid eps of 1, a seed value of 42, and grid update enabled.

\subsubsection{Grid Size}

\begin{figure} [ht]
\begin{center}
\centerline{\includegraphics[width=\columnwidth]{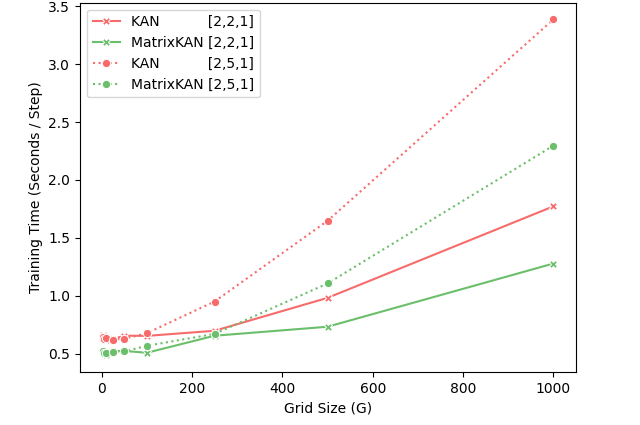}}
\caption{A plot comparing MatrixKAN and KAN models of shape [2,2,1] and [2,5,1] across increasing grid sizes.}
\label{fig:comp_eff_grid_size}
\end{center}
\end{figure}

For our test measuring computational efficiency with respect to grid size,\footnote{As demonstrated in \cite{liu2024}, achieving optimal KAN performance sometimes requires refining the grid to grid sizes of up to 1000. Given the importance of increasing grid size during training to achieve optimal model performance, we included computational efficiency results with respect to increasing grid size to verify that the MatrixKAN optimizations did not create unacceptable additional computational burden at higher grid sizes.} B-spline degree was set to 6, and grid size was varied across the following values: 2, 5, 10, 25, 50, 100, 250, 500, 1000.  Results are depicted in Figure \ref{fig:comp_eff_grid_size} by plotting training time in seconds per training step against grid size.

From Figure \ref{fig:comp_eff_grid_size}, we can see superior performance of MatrixKAN compared to KAN.  As depicted in the figure, as grid size increases, training time for each architecture increases, but MatrixKAN training times are consistently less than that of KAN.
The relatively efficient performance of MatrixKAN vs. KAN in these tests is a result of the static B-spline degree used--6--for which KAN must execute 6 sequential calculations per B-spline that are parallelized by MatrixKAN.  As a result, although both KAN and MatrixKAN demonstrate similar scaling with respect to grid size, as expected from our theoretical analysis, MatrixKAN demonstrates superior performance due to the underlying B-spline degree.

\subsubsection{B-spline Degree}

\begin{figure} [ht]
\begin{center}
\centerline{\includegraphics[width=0.95\columnwidth]{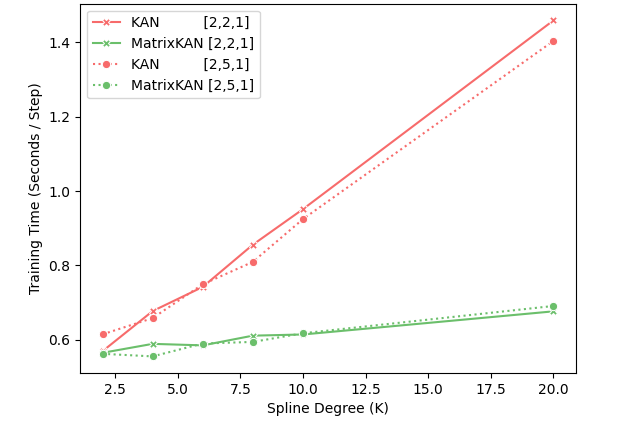}}
\caption{A plot comparing MatrixKAN and KAN models of shape [2,2,1] and [2,5,1] across increasing spline degrees.}
\label{fig:comp_eff_spline_degree}
\end{center}
\end{figure}

For our test measuring computational efficiency with respect to B-spline degree, grid size was set to 2, and B-spline degree was varied across the following values: 2, 4, 6, 8, 10, 20.  Results are depicted in Figure \ref{fig:comp_eff_spline_degree} by plotting training time in seconds per training step against B-spline degree.

\begin{figure*} [!h]
    \centering
    \begin{minipage}{0.32\textwidth}
        \includegraphics[width=\linewidth]{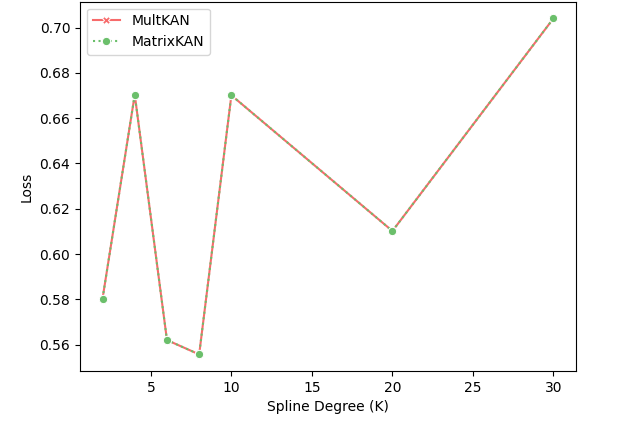}
        \caption{A plot comparing loss level of MatrixKAN and KAN models trained to model Feynman Equation I.6.20b:\\ \(f(\Theta, \sigma)=exp(-\frac{\Theta^2}{2\sigma^2})/\sqrt{2\pi\sigma^2}\).}
        \label{fig:comp_perf_3}
    \end{minipage}
    \hfill
    \begin{minipage}{0.32\textwidth}
        \includegraphics[width=\linewidth]{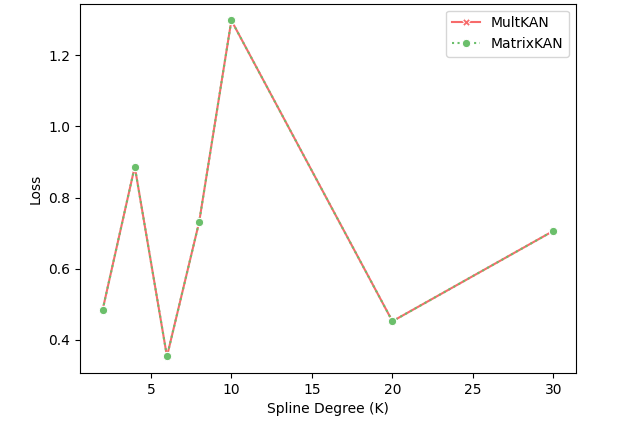}
        \caption{A plot comparing loss level of MatrixKAN and KAN models trained to model Feynman Equation I.12.11:\\ \(f(\alpha, \Theta)=1+\alpha sin\Theta\)}
        \label{fig:comp_perf_12}
    \end{minipage}
    \hfill
    \begin{minipage}{0.32\textwidth}
        \includegraphics[width=\linewidth]{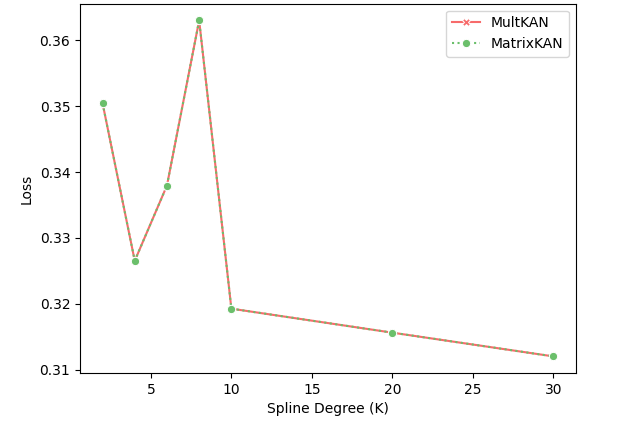}
        \caption{A plot comparing loss level of MatrixKAN and KAN models trained to model Feynman Equation I.26.2:\\ \(f(n, \Theta_2)=arcsin(n sin \Theta_2)\)}
        \label{fig:comp_perf_26}
    \end{minipage}
\end{figure*}

From Figure \ref{fig:comp_eff_spline_degree}, we can see a clear distinction between KAN and MatrixKAN models, demonstrating preferable scaling of MatrixKAN computation time with respect to B-spline degree.  For the base KAN architecture, both model shapes scale as expected, with computation time scaling linearly with increasing B-spline degree.
This supports the conclusion that KAN's effective computation time is a function of B-spline degree and MatrixKAN's is independent of B-spline degree, even though the exact speedup here may not equal to the theoretic value of $k$ (due to implementation variation that may introduce a constant factor of overhead in the asymptotic analysis).

\subsubsection{Dataset Size}

Given that real-world training often involves datasets orders of magnitude larger than those utilized in the tests above, we performed a test to measure the computational efficiency of each architecture with respect to dataset size.  For this test, we trained a model of shape [2,5,1] with grid size set to 3, B-spline degree set to 20, and dataset size varied across the following values: 10000, 50000, 100000.  Results are depicted in Table \ref{table:speedups} comparing dataset size to relative speedup, calculated as seconds per step for KAN calculations divided by seconds per step for MatrixKAN calculations.

\begin{table}[ht]
\caption{Speedups of MatrixKAN vs. KAN for various dataset sizes.}
\label{table:speedups}
\vskip 0.15in
\begin{center}
\begin{small}
\begin{sc}
\begin{tabular}{lcccc}
\toprule
    \textbf{Dataset Size} & 10000 & 50000 & 100000 \\
\midrule
    \textbf{Speedup} & 2.18 & 17.59 & 38.89 \\
\bottomrule
\end{tabular}
\end{sc}
\end{small}
\end{center}
\vskip -0.1in
\end{table}
\vspace{4mm}

As depicted in the table, as dataset size increases, the relative speedup of MatrixKAN vs. KAN increases significantly. For datasets up to 100,000 samples, MatrixKAN computations exhibit speedups of up to approximately 40x over KAN computations. 
The increased speedup is mainly because the parallel fraction of MatrixKAN is greater than that of KAN, so the proportion of computations performed in parallel by MatrixKAN increases more quickly than KAN over increasing dataset sizes. Therefore, as dataset size increases, MatrixKAN's speedups relative to KAN will increase, with potential speedups many times, if not orders of magnitude, greater than those depicted here for large dataset sizes.

\subsection{Performance / Optimal Spline Degree}

For our performance tests, we compared the performance of MatrixKAN to KAN by plotting the loss level of each model against B-spline degree. The results of select tests are shown in Figures \ref{fig:comp_perf_3}, \ref{fig:comp_perf_12}, and \ref{fig:comp_perf_26}.

Two things can be observed. First, MatrixKAN and KAN performed identically modeling all Feynman equations (i.e., completely overlapped curves).  Although MatrixKAN and KAN perform B-spline calculations in different ways, the calculations are identical.  Thus, when initialized with identical parameters and trained on identical datasets, models from each architecture should produce identical outputs and follow the same path of gradient descent to identical levels of convergence, which is demonstrated here.  Second, the results demonstrate the superior performance of high-degree B-splines (i.e. of degree 4 or higher) in modeling various functions.  Of the nine Feynman functions modeled, all but two models achieved optimal convergence at B-spline degree 4 or higher, with the majority of models converging at B-spline degree 6 or higher with loss level improvements of up to 27.0\%.  Further, Figure \ref{fig:comp_perf_26} demonstrates that for some modeling tasks model performance generally improves as a function of increasing B-spline degree and that B-splines of higher degree than tested here may yield improved results.  

\section{Conclusion}

Kolmogorov-Arnold Networks represent a promising innovation in neural network architecture.  However, due to the recursive nature of B-spline calculations, KANs suffer from slow training and inference speeds relative to traditional MLPs.  To address this issue, we propose MatrixKAN, a novel optimization that parallelizes B-spline calculations.  This optimized architecture adapts the generalized matrix representation of B-splines to parallelize B-spline calculations, resulting in significant speedups relative to KAN.  Using this optimized approach, we observed speedups of up to approximately 40x compared to KAN, with potential for significant additional speedups for larger datasets or higher B-spline degrees, while demonstrating consistent model performance between architectures.

\pagebreak
\section*{Impact Statement}

This paper presents work whose goal is to advance the field of Machine Learning. It demonstrates various potential societal benefits, in particular those resulting from the increased testing and use of KAN models using high-degree B-splines.  There are many potential societal consequences of our work, none which we feel must be specifically highlighted here.

% In the unusual situation where you want a paper to appear in the
% references without citing it in the main text, use \nocite
% \nocite{langley00}

\bibliography{MatrixKAN}

\begin{thebibliography}{23}
\providecommand{\natexlab}[1]{#1}
\providecommand{\url}[1]{\texttt{#1}}
\expandafter\ifx\csname urlstyle\endcsname\relax
  \providecommand{\doi}[1]{doi: #1}\else
  \providecommand{\doi}{doi: \begingroup \urlstyle{rm}\Url}\fi

\bibitem[Aghaei(2024)]{aghaei2024}
Aghaei, A.~A.
\newblock rkan: Rational kolmogorov-arnold networks, 2024.
\newblock URL \url{https://arxiv.org/abs/2406.14495}.

\bibitem[Blelloch(1996)]{blelloch1996}
Blelloch, G.~E.
\newblock Programming parallel algorithms.
\newblock \emph{Commun. ACM}, 39\penalty0 (3):\penalty0 85–97, March 1996.
\newblock ISSN 0001-0782.
\newblock \doi{10.1145/227234.227246}.
\newblock URL \url{https://doi.org/10.1145/227234.227246}.

\bibitem[Bodner et~al.(2024)Bodner, Tepsich, Spolski, and Pourteau]{bodner2024}
Bodner, A.~D., Tepsich, A.~S., Spolski, J.~N., and Pourteau, S.
\newblock Convolutional kolmogorov-arnold networks, 2024.
\newblock URL \url{https://arxiv.org/abs/2406.13155}.

\bibitem[Bozorgasl \& Chen(2024)Bozorgasl and Chen]{bozorgasl2024}
Bozorgasl, Z. and Chen, H.
\newblock Wav-kan: Wavelet kolmogorov-arnold networks, 2024.
\newblock URL \url{https://arxiv.org/abs/2405.12832}.

\bibitem[Carlo et~al.(2024)Carlo, Mastropietro, and Anagnostopoulos]{decarlo2024}
Carlo, G.~D., Mastropietro, A., and Anagnostopoulos, A.
\newblock Kolmogorov-arnold graph neural networks, 2024.
\newblock URL \url{https://arxiv.org/abs/2406.18354}.

\bibitem[Chen \& Zhang(2024{\natexlab{a}})Chen and Zhang]{chen2024larctan}
Chen, Z. and Zhang, X.
\newblock Larctan-skan: Simple and efficient single-parameterized kolmogorov-arnold networks using learnable trigonometric function, 2024{\natexlab{a}}.
\newblock URL \url{https://arxiv.org/abs/2410.19360}.

\bibitem[Chen \& Zhang(2024{\natexlab{b}})Chen and Zhang]{chen2024lss}
Chen, Z. and Zhang, X.
\newblock Lss-skan: Efficient kolmogorov-arnold networks based on single-parameterized function, 2024{\natexlab{b}}.
\newblock URL \url{https://arxiv.org/abs/2410.14951}.

\bibitem[Cohen \& Riesenfeld(1982)Cohen and Riesenfeld]{cohen1982}
Cohen, E. and Riesenfeld, R.~F.
\newblock General matrix representations for bezier and b-spline curves.
\newblock \emph{Computers in Industry}, 3\penalty0 (1):\penalty0 9--15, 1982.
\newblock ISSN 0166-3615.
\newblock \doi{https://doi.org/10.1016/0166-3615(82)90027-6}.
\newblock URL \url{https://www.sciencedirect.com/science/article/pii/0166361582900276}.
\newblock Double Issue- In Memory of Steven Anson Coons.

\bibitem[Hu et~al.(2024)Hu, Wang, and Lin]{hu2024}
Hu, L., Wang, Y., and Lin, Z.
\newblock Incorporating arbitrary matrix group equivariance into kans, 2024.
\newblock URL \url{https://arxiv.org/abs/2410.00435}.

\bibitem[Jamali et~al.(2024)Jamali, Roy, Hong, Lu, and Ghamisi]{jamali2024}
Jamali, A., Roy, S.~K., Hong, D., Lu, B., and Ghamisi, P.
\newblock How to learn more? exploring kolmogorov-arnold networks for hyperspectral image classification, 2024.
\newblock URL \url{https://arxiv.org/abs/2406.15719}.

\bibitem[Koenig et~al.(2024)Koenig, Kim, and Deng]{Koenig2024}
Koenig, B.~C., Kim, S., and Deng, S.
\newblock Kan-odes: Kolmogorov–arnold network ordinary differential equations for learning dynamical systems and hidden physics.
\newblock \emph{Computer Methods in Applied Mechanics and Engineering}, 432:\penalty0 117397, December 2024.
\newblock ISSN 0045-7825.
\newblock \doi{10.1016/j.cma.2024.117397}.
\newblock URL \url{http://dx.doi.org/10.1016/j.cma.2024.117397}.

\bibitem[Kolmogorov(1957)]{kolmogorov1957}
Kolmogorov, A.~K.
\newblock On the representation of continuous functions of several variables by superposition of continuous functions of one variable and addition.
\newblock \emph{Doklady Akademii Nauk SSSR}, 114:\penalty0 369--373, 1957.

\bibitem[Li et~al.(2024)Li, Liu, Li, Wang, Liu, Liu, Chen, and Yuan]{li2024}
Li, C., Liu, X., Li, W., Wang, C., Liu, H., Liu, Y., Chen, Z., and Yuan, Y.
\newblock U-kan makes strong backbone for medical image segmentation and generation, 2024.
\newblock URL \url{https://arxiv.org/abs/2406.02918}.

\bibitem[Li(2024)]{li2024radial}
Li, Z.
\newblock Kolmogorov-arnold networks are radial basis function networks, 2024.
\newblock URL \url{https://arxiv.org/abs/2405.06721}.

\bibitem[Liu et~al.(2024{\natexlab{a}})Liu, Wang, Vaidya, Ruehle, Halverson, Soljačić, Hou, and Tegmark]{liu2024}
Liu, Z., Wang, Y., Vaidya, S., Ruehle, F., Halverson, J., Soljačić, M., Hou, T.~Y., and Tegmark, M.
\newblock Kan: Kolmogorov-arnold networks, 2024{\natexlab{a}}.
\newblock URL \url{https://arxiv.org/abs/2404.19756}.

\bibitem[Liu et~al.(2024{\natexlab{b}})Liu, Wang, Vaidya, Ruehle, Halverson, Soljačić, Hou, and Tegmark]{pykan}
Liu, Z., Wang, Y., Vaidya, S., Ruehle, F., Halverson, J., Soljačić, M., Hou, T.~Y., and Tegmark, M.
\newblock pykan.
\newblock \url{https://github.com/kindxiaoming/pykan}, 2024{\natexlab{b}}.
\newblock Accessed: January 28, 2025.

\bibitem[Liu et~al.(2025)Liu, Chen, Bai, Li, Zou, and Shi]{liuchen2025}
Liu, Z., Chen, H., Bai, L., Li, W., Zou, Z., and Shi, Z.
\newblock Kolmogorov arnold neural interpolator for downscaling and correcting meteorological fields from in-situ observations, 2025.
\newblock URL \url{https://arxiv.org/abs/2501.14404}.

\bibitem[Peng et~al.(2024)Peng, Wang, Hu, He, Mao, Huang, and Ding]{peng2024}
Peng, Y., Wang, Y., Hu, F., He, M., Mao, Z., Huang, X., and Ding, J.
\newblock Predictive modeling of flexible ehd pumps using kolmogorov–arnold networks.
\newblock \emph{Biomimetic Intelligence and Robotics}, 4\penalty0 (4):\penalty0 100184, December 2024.
\newblock ISSN 2667-3797.
\newblock \doi{10.1016/j.birob.2024.100184}.
\newblock URL \url{http://dx.doi.org/10.1016/j.birob.2024.100184}.

\bibitem[Prautzsch et~al.(2002)Prautzsch, Boehm, and Paluszny]{prautzsch2002}
Prautzsch, H., Boehm, W., and Paluszny, M.
\newblock \emph{Bézier and B-Spline Techniques}.
\newblock Springer Berlin, Heidelberg, 01 2002.
\newblock ISBN 978-3-642-07842-2.
\newblock \doi{10.1007/978-3-662-04919-8}.

\bibitem[Qin(1998)]{qin1998}
Qin, K.
\newblock General matrix representations for b-splines.
\newblock \emph{The Visual Computer}, 16:\penalty0 177--186, 1998.
\newblock URL \url{https://api.semanticscholar.org/CorpusID:26401557}.

\bibitem[Samadi et~al.(2024)Samadi, Müller, and Schuppert]{samadi2024}
Samadi, M.~E., Müller, Y., and Schuppert, A.
\newblock Smooth kolmogorov arnold networks enabling structural knowledge representation, 2024.
\newblock URL \url{https://arxiv.org/abs/2405.11318}.

\bibitem[Zheng et~al.(2025)Zheng, Zhang, Yue, Xu, Maennel, and Chen]{zheng2025}
Zheng, L.~N., Zhang, W.~E., Yue, L., Xu, M., Maennel, O., and Chen, W.
\newblock Free-knots kolmogorov-arnold network: On the analysis of spline knots and advancing stability, 2025.
\newblock URL \url{https://arxiv.org/abs/2501.09283}.

\bibitem[Zhou et~al.(2024)Zhou, Pei, Sun, Han, Gao, Pei, Zhang, Xie, and Li]{zhou2024}
Zhou, Q., Pei, C., Sun, F., Han, J., Gao, Z., Pei, D., Zhang, H., Xie, G., and Li, J.
\newblock Kan-ad: Time series anomaly detection with kolmogorov-arnold networks, 2024.
\newblock URL \url{https://arxiv.org/abs/2411.00278}.

\end{thebibliography}
\bibliographystyle{icml2025}

%%%%%%%%%%%%%%%%%%%%%%%%%%%%%%%%%%%%%%%%%%%%%%%%%%%%%%%%%%%%%%%%%%%%%%%%%%%%%%%
%%%%%%%%%%%%%%%%%%%%%%%%%%%%%%%%%%%%%%%%%%%%%%%%%%%%%%%%%%%%%%%%%%%%%%%%%%%%%%%
% APPENDIX
%%%%%%%%%%%%%%%%%%%%%%%%%%%%%%%%%%%%%%%%%%%%%%%%%%%%%%%%%%%%%%%%%%%%%%%%%%%%%%%
%%%%%%%%%%%%%%%%%%%%%%%%%%%%%%%%%%%%%%%%%%%%%%%%%%%%%%%%%%%%%%%%%%%%%%%%%%%%%%%
\newpage
\appendix
\onecolumn
\section{Representing the product of two polynomials using Toeplitz matrices.} \label{Toeplitz}

A Toeplitz matrix  is a matrix whose elements on any line parallel to the main diagonal are all equal:

\[
T =
\begin{bmatrix}
    \alpha_0 & \alpha_1 & ... & \alpha_s &  ... & 0 \\
    \alpha_{-1} & \alpha_0 & \ddots &  & \ddots &  \\
    \vdots & \ddots & \ddots & \ddots & \ddots & \alpha_s \\
    \alpha_{-r} & \ddots & \ddots & \ddots & \ddots & \vdots \\
    \vdots & \ddots & \ddots & \ddots & \ddots & \alpha_1 \\
    0 &  & \alpha_{-r} & \ddots & \alpha_{-1} & \alpha_0  \\
\end{bmatrix}
\]

Toeplitz matrices are capable of representing polynomial functions as well as the product of polynomial functions.  For example, the polynomial \(g(x) = c_0 + c_1x + c_2x^2 + ... + c_{m-1}x^{m-1} (c_{m-1} \ne 0)\) may be represented as a special Toeplitz matrix—a lower triangular matrix—as follows:

\[g(x) = 
\begin{bmatrix}
    c_0 & & & & & & 0 \\
    c_1 & c_0 & & & & & \\
    \vdots & \ddots & \ddots & & & & \\
    c_{m-1} & ... & \ddots & \ddots & & & \\
     & \ddots & ... & \ddots & \ddots & & \\
     & & c_{m-1} & ... & \ddots & c_0 & \\
     0 & & & c_{m-1} & ... & c_1 & c_0
\end{bmatrix}
\]

Further, the product of \(g(x)\) and \(q(x) = d_0 + d_1x + d_2x^2 + ... + d_{n-1}x^{n-1} (d_{n-1} \ne 0)\) can be represented as follows:

\[
\begin{aligned}
    f(x) & = g(x)q(x) \\
    & = X
    \begin{bmatrix}
        c_0 & & & & & & 0 \\
        c_1 & c_0 & & & & & \\
        \vdots & \ddots & \ddots & & & & \\
        c_{m-1} & ... & \ddots & \ddots & & & \\
         & \ddots & ... & \ddots & \ddots & & \\
         & & c_{m-1} & ... & \ddots & c_0 & \\
         0 & & & c_{m-1} & ... & c_1 & c_0
    \end{bmatrix}
    \begin{bmatrix}
        d_0 \\
        d_1 \\
        \vdots \\
        d_{n-1} \\
        0 \\
        \vdots \\
        0
    \end{bmatrix}
\end{aligned}
\]

where \(X =
\begin{bmatrix}
    1 & x & x^2 & ... & x^{m+n-2}
\end{bmatrix}\).

%%%%%%%%%%%%%%%%%%%%%%%%%%%%%%%%%%%%%%%%%%%%%%%%%%%%%%%%%%%%%%%%%%%%%%%%%%%%%%%
%%%%%%%%%%%%%%%%%%%%%%%%%%%%%%%%%%%%%%%%%%%%%%%%%%%%%%%%%%%%%%%%%%%%%%%%%%%%%%%

\end{document}